\documentclass[12pt]{article}

\usepackage{epstopdf}
\usepackage{float}
\usepackage{units}
\usepackage{amsmath}
\usepackage{amssymb}
\usepackage{graphicx}

\floatstyle{ruled}
\newfloat{algorithm}{tbp}{loa}
\providecommand{\algorithmname}{Algorithm}
\floatname{algorithm}{\protect\algorithmname}

\begin{document}

\title{Global hard thresholding algorithms for joint sparse image representation and denoising}

\author{Reza Borhani\and
        Jeremy Watt\and
        Aggelos Katsaggelos
}
\date{}
\maketitle






\begin{abstract}
Sparse coding of images is traditionally done by cutting them into small patches and representing each patch individually over some dictionary given a pre-determined number of nonzero coefficients to use for each patch. In lack of a way to effectively distribute a total number (or global budget) of nonzero coefficients across all patches, current sparse recovery algorithms distribute the global budget equally across all patches despite the wide range of differences in structural complexity among them. In this work we propose a new framework for joint sparse representation and recovery of all image patches simultaneously. We also present two novel global hard thresholding algorithms, based on the notion of variable splitting, for solving the joint sparse model. Experimentation using both synthetic and real data shows effectiveness of the proposed framework for sparse image representation and denoising tasks. Additionally, time complexity analysis of the proposed algorithms indicate high scalability of both algorithms, making them favorable to use on large megapixel images.
\end{abstract}


\section{Introduction}

\label{sec:intro}

In recent years a large number of algorithms have been developed for
approximately solving the NP-hard sparse representation problem

\noindent
\begin{equation}
\begin{aligned}\underset{\mathbf{x}}{\mbox{minimize}} & \,\,\Vert\mathbf{D}\mathbf{x}-\mathbf{y}\Vert_{2}^{2}\\
\mbox{subject to} & \,\,\Vert\mathbf{x}\Vert_{0}\leq s,
\end{aligned}
\label{eq:Standard L-0 problem}
\end{equation}

\noindent where $\mathbf{y}$ is a signal of dimension $N\times1$,
$\mathbf{D}$ is an $N\times L$ dictionary, $\mathbf{x}$ is an $L\times1$
coefficient vector, and $\Vert\cdot\Vert_{0}$ denotes the $\ell_{0}$
norm that counts the number of nonzero entries in a vector (or matrix).
These approaches can be roughly divided into three categories. First,
\textit{greedy pursuit approaches} such as the popular Orthogonal Matching
Pursuit (OMP) algorithm \cite{pati1993orthogonal,tropp2007signal}
which sequentially adds new atoms (or dictionary elements) to a signal
representation in a greedy fashion until the entire budget of $s$
nonzero coefficients is used. Second, \textit{convex relaxation approaches}
like the Fast Iterative Shrinkage Thresholding Algorithm (FISTA) \cite{beck2009fast}
wherein the $\ell_{0}$ norm of the coefficient vector $\mathbf{x}$
is appropriately weighted, brought up to the objective function, and
replaced with an $\ell_{1}$ norm. This convex relaxation of the sparse
representation problem in (\ref{eq:Standard L-0 problem}) is then
solved via accelerated proximal gradient. Lastly, \textit{hard thresholding
approaches} such as the Accelerated Iterative Hard Thresholding (AIHT)
algorithm \cite{blumensath2012accelerated} which approximately solves
(\ref{eq:Standard L-0 problem}) by projected gradient descent onto
the nonconvex set of $s$-sparse vectors given by $\left\{ \mathbf{x}\in\mathbb{R}^{L}\,\vert\,\Vert\mathbf{x}\Vert_{0}\leq s\right\} $.

\begin{figure*}[!th]
\noindent \begin{centering}
\includegraphics[scale=0.27]{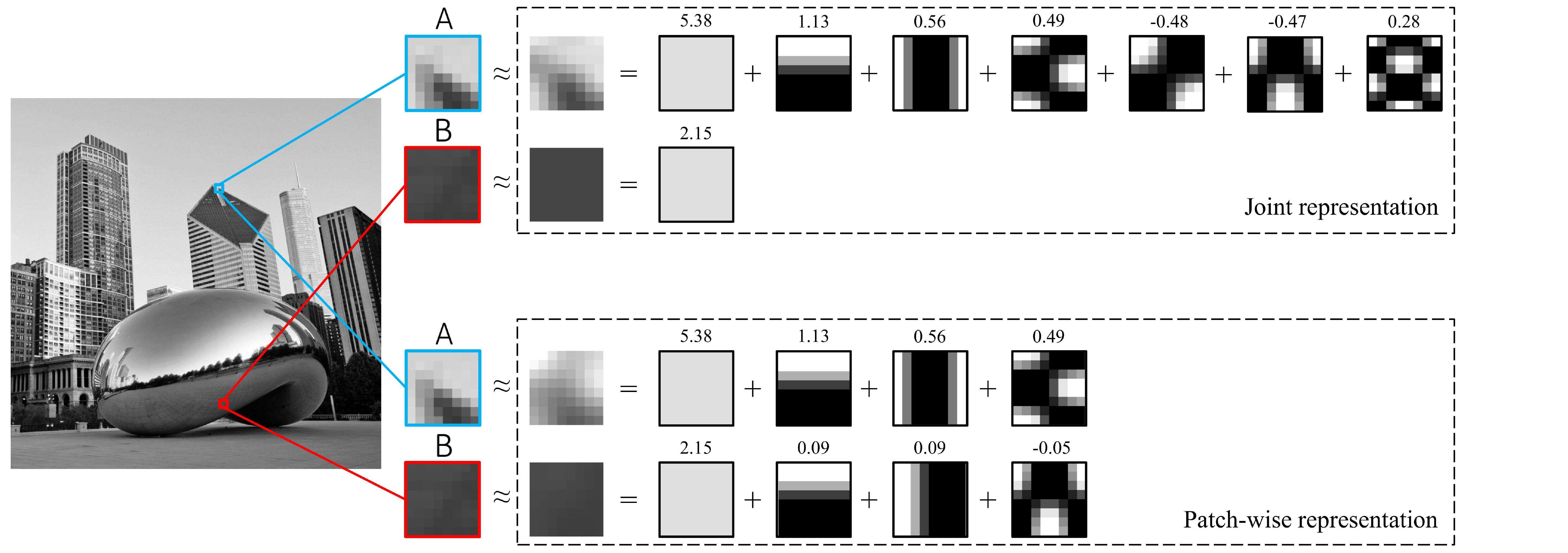}
\par\end{centering}

\caption{\label{fig:gloval_vs_patch}(top) Joint and (bottom) patch-wise representations
of two $8\times8$ patches using a total budget of $8$ DCT atoms,
where the respective coefficient is written on top of each atom. Comparing
the reconstructed patches in the bottom panel to those in the top panel
shows an intangible improvement gained by adding more atoms to the
representation of patch B at the
expense of a significant deterioration in reconstructing patch A. This simple observation justifies
the intuitive choice of allocating a larger number of atoms to represent
patches with more complex structure, as well as algorithms for automatically
making smart global budget allocations. }
\end{figure*}

Greedy pursuit and convex relaxation approaches have received significant
attention from researchers in signal and image processing (see e.g.,
\cite{tropp2006algorithms,tropp2007signal,pati1993orthogonal,beck2009fast,combettes2005signal}).
However, several recent works have suggested that hard thresholding
routines not only have strong recovery guarantees, but in practice
can outperform greedy pursuit or convex relaxation approaches, particularly
in compressive sensing \cite{candes2006robust,donoho2006cs} applications, both in terms of efficacy and
computation time \cite{blumensath2012accelerated,blumensath2010normalized,garg2009gradient}.
Current hard thresholding routines that are proposed to solve the
general constrained optimization problem

\noindent
\begin{equation}
\begin{aligned}\underset{\mathbf{x}}{\mbox{minimize}} & \,\, f\left(\mathbf{x}\right)\\
\mbox{subject to} & \,\,\,\mathbf{x}\in\mathcal{T},
\end{aligned}
\label{eq:Standard L-0 problem-1}
\end{equation}

\noindent are largely based on the projected gradient method, wherein
at the $k^{th}$ iteration a gradient descent step is taken in the
objective function $f$ and is then adjusted via an appropriate operator
as
\begin{equation}
\mathbf{x}^{k}=P_{\mathcal{T}}\left(\mathbf{x}^{k-1}-\alpha_{k}\nabla f\left(\mathbf{x}^{k-1}\right)\right).
\end{equation}

\noindent Here the gradient step $\mathbf{x}^{k-1}-\alpha_{k}\nabla f\left(\mathbf{x}^{k-1}\right)$,
where $\alpha_{k}$ is an appropriately chosen step-length, is transformed
by the projection operator $P_{\mathcal{T}}\left(\cdot\right)$ so
that each step in the procedure remains in the problem's given constraint
set $\mathcal{T}$. Denoting by $\mathcal{H}_{s}\left(\cdot\right)$
the projection onto the $s$-sparse set, we have
\begin{equation}
\begin{aligned}\mathcal{H}_{s}\left(\mathbf{y}\right)=\underset{\left\Vert \mathbf{x}\right\Vert _{0}\leq s}{\mbox{argmin}} & \,\,\Vert\mathbf{x}-\mathbf{y}\Vert_{2}^{2}\end{aligned}
\end{equation}
where $\mathcal{H}_{s}\left(\mathbf{y}\right)$ is a vector of equal
length to $\mathbf{y}$ wherein we keep only the top $s$ (in magnitude)
entries of $\mathbf{y}$, and set the remaining elements to zero.
With this notation, popular hard thresholding approaches \cite{blumensath2009iterative,blumensath2010normalized,blumensath2012accelerated,garg2009gradient}
for solving (\ref{eq:Standard L-0 problem}) take projected gradient
steps of the form
\begin{equation}
\mathbf{x}^{k}=\mathcal{H}_{s}\left(\mathbf{x}^{k-1}-\alpha_{k}\mathbf{D}^{T}\left(\mathbf{D}\mathbf{x}^{k-1}-\mathbf{y}\right)\right).
\end{equation}
Since the $s$-sparse set is nonconvex one might not expect projected
gradient to converge at all, let alone to a sufficiently low objective
value. However projected gradient in this instance is in fact provably
convergent when the dictionary $\mathbf{D}$ satisfies various forms
of a Restricted Isometry Property (RIP) \cite{candes2005decode}, i.e., if $\mathbf{D}$ satisfies
\begin{equation}
\left(1-\delta_{s}\right)\left\Vert \mathbf{x}\right\Vert _{2}^{2}\leq\left\Vert \mathbf{D}\mathbf{x}\right\Vert _{2}^{2}\leq\left(1+\delta_{s}\right)\left\Vert \mathbf{x}\right\Vert _{2}^{2}
\end{equation}
for all $s$-sparse vectors $\mathbf{x}$ and for some $\delta_{s}\in\left(0,1\right)$.
Such a matrix is used almost exclusively in compressive sensing applications.
Analogous projected gradient methods have been successfully applied
to the low-rank matrix completion problem \cite{jain2010guaranteed},
where hard thresholding is performed on singular values as opposed
to entries of a matrix itself, and has also been shown to be theoretically
and practically superior to standard convex relaxation approaches
which invoke the rank-convexifying surrogate, the nuclear norm \cite{meka2009guaranteed},
when RIP conditions hold for the problem. It must be noted that while
these algorithms have mathematically guaranteed convergence for RIP-based
problems, it is unclear how well they contend on the plethora of other
instances of (\ref{eq:Standard L-0 problem}) where the matrix $\mathbf{D}$
does not necessarily hold an RIP (e.g., image denoising and deblurring \cite{elad2006image,elad2010role}, super-resolution \cite{yang2010image,candes2014superres}, sparse coding \cite{olshausen1997sparse}).

\subsection{Joint sparse representation model}

\begin{figure*}
\noindent \begin{centering}
\includegraphics[scale=0.32]{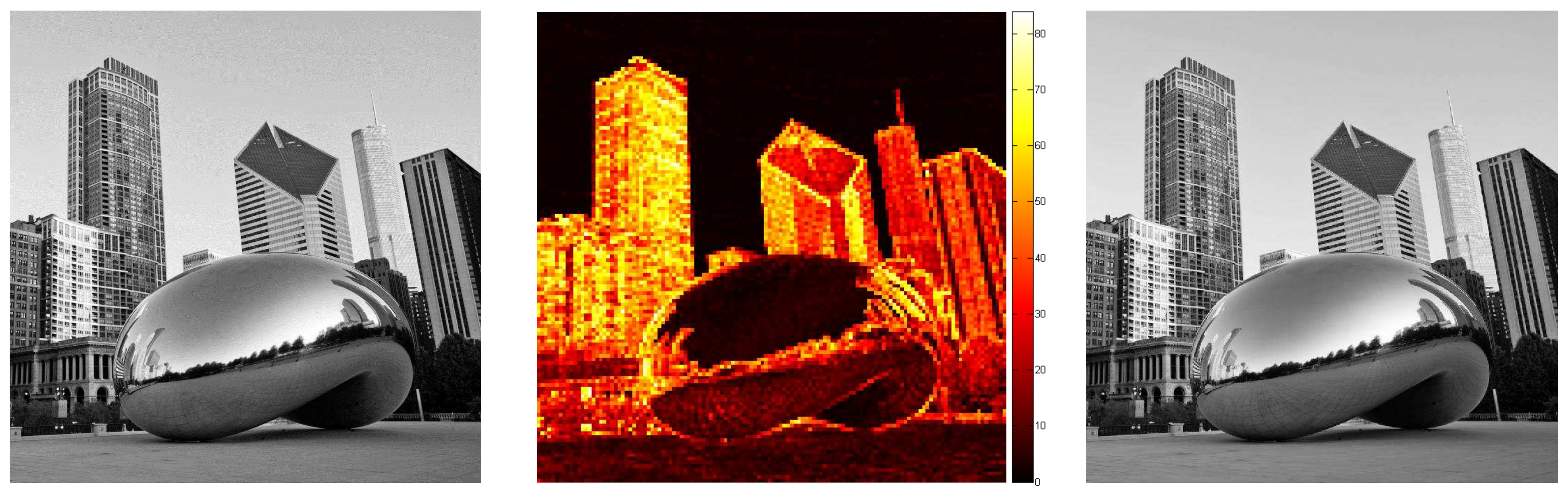}
\par\end{centering}

\caption{\label{fig:heatmap }With an algorithm for solving the global sparse
representation problem we can automatically distribute a global budget
where it really needs to be: on high frequency areas of an image.
(left) The original image. (middle) The heatmap generated by counting the
number of atoms used in reconstructing each patch. (right) The reconstructed
image with a root mean square error of less than $1$. }
\end{figure*}

In many image processing applications large or even moderate sized
images are cut into small image patches (or blocks), and then one
wants to sparsely represent a large number of patches $\left\{ \mathbf{y}_{p}\right\} _{p=1}^{P}$
together, given a \emph{global budget }$S$ for the total number of
nonzero coefficients to use. This ideally requires the user to decide
on the individual per patch budget $s_{p}$ for each of the $P$ patches
in a way to ensure that $\underset{p}{\sum}s_{p}\leq S$. Because
this global budget allocation problem seems difficult to solve, in
practice a fixed $s=\lfloor\frac{S}{P}\rfloor$ is typically chosen
for all patches, even though this choice results in a suboptimal distribution
of the global budget considering the wide range of differences in
structural complexity across the patches. This is particularly the
case with natural images wherein patches vary extremely in terms of
texture, structure, and frequency content. We illustrate this observation
through a simple example in Figure \ref{fig:gloval_vs_patch} where
two $8\times8$ patches taken from an image are sparsely represented
over the $64\times64$ Discrete Cosine Transform (DCT) dictionary.
One of the patches (patch B in Figure \ref{fig:gloval_vs_patch}) is rather flat and can
be represented quite well using only one atom from the DCT dictionary
while the other more structurally complex patch (patch A) requires at least $7$ atoms in order to be represented equally
well in terms of reconstruction error. Notice that the naive way of
distributing a total of $8$ nonzero coefficients equally across both
patches ($4$ atoms per patch) would adversely affect the representation
of the more complex patch with no tangible improvement in the representation
of the flatter one.

This observation motivates introduction of the joint sparse representation
problem, where local patch-wise budgets can be determined automatically
via solving

\noindent
\begin{equation}
\begin{aligned}\underset{\mathbf{X}}{\mbox{minimize}} & \,\,\Vert\mathbf{D}\mathbf{X}-\mathbf{Y}\Vert_{F}^{2}\\
\mbox{subject to} & \,\,\Vert\mathbf{X}\Vert_{0}\leq S,
\end{aligned}
\label{eq: Simultaneous L0 problem}
\end{equation}

\noindent where we have concatenated all signals $\left\{ \mathbf{y}_{p}\right\} _{p=1}^{P}$
into an $N\times P$ matrix $\mathbf{Y}$, $\mathbf{X}$ is the corresponding
coefficient matrix of size $L\times P$, and $\Vert\cdot\Vert_{F}$ denotes
the Frobenius norm. If this problem could be solved efficiently, the
issue of how to distribute the budget $S$ across all $P$ patches
would be taken care of automatically, alleviating the painstaking
per patch budget tuning required when applying (\ref{eq:Standard L-0 problem})
to each individual patch. Note that one could concatenate all columns in the matrix $\mathbf{X}$ into a single vector and then use any of the patch-wise algorithms designed for solving (\ref{eq:Standard L-0 problem}). This solution however is not practically feasible due to the potentially
large size of $\mathbf{X}$.

\subsection{Proposed approaches}

The hard thresholding approaches described in this work for solving (\ref{eq: Simultaneous L0 problem}) are based
on the notion of variable splitting as well as two classic approaches
to constrained numerical optimization. More specifically, in this
work we present two scalable hard thresholding approaches for approximately
solving the joint sparse representation problem in (\ref{eq: Simultaneous L0 problem}).
The first approach, based on variable splitting and the Quadratic
Penalty Method (QPM) \cite{nocedal2006penalty}, is a provably convergent
method while the latter employs a heuristic form of the Alternating
Direction Method of Multipliers (ADMM) framework \cite{boyd2011distributed}.
While ADMM is often applied to convex optimization problems (where
it is provably convergent), our experiments add to the growing body
of work showing that ADMM can be a highly effective empirical heuristic
method for nonconvex optimization problems.

To illustrate what can be achieved by solving the joint model
in (\ref{eq: Simultaneous L0 problem}), we show in Figure \ref{fig:heatmap }
the result of applying our first global hard thresholding algorithm
to sparsely represent a megapixel image. Specifically, we sparsely
represent a gray-scale image of size $1024\times1024$ over a $64\times100$
overcomplete Discrete Cosine Transform (DCT) dictionary using a fixed
global budget of $S=20\times P$, where $P$ is the number of non-overlapping
$8\times8$ patches which constitute the original image. We also
keep count of the number of atoms used in reconstructing each patch,
that is the count $\left\Vert \mathbf{x}_{p}\right\Vert _{0}$ of
nonzero coefficients in the final representation $\mathbf{D}\mathbf{x}_{p}$
for each patch $\mathbf{y}_{p}$, and form a heatmap of the same size
as the original image in order to provide initial visual verification
that our algorithm properly distributes the global budget. In the heatmap the brighter the patch color the more atoms are assigned
in reconstructing it. As can be seen in the middle panel of Figure
\ref{fig:heatmap }, our algorithm appears to properly allocate fractions
of the budget to high frequency portions of the image.

The remainder of this work is organized as follows: in the next Section
we derive both Global Hard Thresholding (GHT) algorithms, referred
to as GHT-QPM and GHT-ADMM hereafter, followed by a complete time
complexity analysis of each algorithm. Then in Section \ref{sec:Experiments} we discuss
the experimental results of applying both algorithms to sparse image representation and denoising tasks. Finally we conclude this paper in
Section \ref{sec:Conclusions} with reflections and thoughts on future work.

\section{Global Hard Thresholding}
\label{sec:Global-Hard-Thresholding}

In this work we introduce two new hard thresholding algorithms that
are effectively applied to the joint sparse representation problem
in (\ref{eq: Simultaneous L0 problem}). Both methods are based on
variable-splitting, as opposed to the projected gradient technique,
and unlike the methods discussed in Section \ref{sec:intro} do
not rely on any kind of RIP condition on the dictionary matrix $\mathbf{D}$.

\subsection{GHT-QPM \label{sub:GHT-QPM}}

The first method we introduce is based on variable splitting and the
Quadratic Penalty Method (QPM). By splitting the optimization variable
we may equivalently rewrite the joint sparse representation problem
from equation (\ref{eq: Simultaneous L0 problem}) as

\noindent
\begin{equation}
\begin{aligned}\underset{\mathbf{X},\mathbf{Z}}{\mbox{minimize}} & \,\,\Vert\mathbf{D}\mathbf{X}-\mathbf{Y}\Vert_{F}^{2}\\
\mbox{subject to} & \,\,\Vert\mathbf{Z}\Vert_{0}\leq S\\
 & \,\,\mathbf{X}=\mathbf{Z}.
\end{aligned}
\end{equation}

\noindent Using QPM we may relax this version of the problem by bringing
the equality constraint to the objective in weighted and squared norm
as

\noindent
\begin{equation}
\begin{aligned}\underset{\mathbf{X},\mathbf{Z}}{\mbox{minimize}} & \,\,\Vert\mathbf{D}\mathbf{X}-\mathbf{Y}\Vert_{F}^{2}+\rho\Vert\mathbf{X}-\mathbf{Z}\Vert_{F}^{2}\\
\mbox{subject to} & \,\,\Vert\mathbf{Z}\Vert_{0}\leq S,
\end{aligned}
\label{eq:relaxed version}
\end{equation}

\noindent where $\rho>0$ controls how well the equality constraint
holds. A simple alternating minimization approach can then be applied
to solving this relaxed form of the joint problem. Specifically, at
the $k^{th}$ step we solve for the following two closed form update
steps first by minimizing the objective of (\ref{eq:relaxed version})
with respect to $\mathbf{X}$ with $\mathbf{Z}$ fixed at its previous
value $\mathbf{Z}^{k-1}$, as

\noindent
\begin{equation}
\mathbf{X}^{k}=\underset{\mathbf{X}}{\mbox{argmin}}\Vert\mathbf{D}\mathbf{X}-\mathbf{Y}\Vert_{F}^{2}+\rho\Vert\mathbf{X}-\mathbf{Z}^{k-1}\Vert_{F}^{2},
\end{equation}

\noindent which can be written in closed form as the solution to the
linear system

\noindent
\begin{equation}
\left(\mathbf{D}^{T}\mathbf{D}+\rho\mathbf{I}\right)\mathbf{X}=\mathbf{D}^{T}\mathbf{Y}+\rho\mathbf{Z}^{k-1},
\end{equation}

\noindent and can be solved for in closed form as

\noindent
\begin{equation}
\mathbf{X}^{k}=\left(\mathbf{D}^{T}\mathbf{D}+\rho\mathbf{I}\right)^{-1}\left(\mathbf{D}^{T}\mathbf{Y}+\rho\mathbf{Z}^{k-1}\right).
\end{equation}

\noindent However we note that in practice such a linear system is
almost never solved by actually inverting the matrix $\mathbf{D}^{T}\mathbf{D}+\rho\mathbf{I}$,
since solving the linear system directly via numerical linear algebra
methods is significantly more efficient. Moreover, since in our case
this matrix remains unchanged throughout the iterations significant
additional computation savings can be achieved by catching a Cholesky
factorization of $\mathbf{D}^{T}\mathbf{D}+\rho\mathbf{I}$. We discuss
this further in Subsection \ref{sub:Complexity-analysis}.

Next, minimizing the objective of (\ref{eq:relaxed version}) with
respect to $\mathbf{Z}$ gives the projection problem

\noindent
\begin{equation}
\mathbf{Z}^{k}=\underset{\left\Vert \mathbf{Z}\right\Vert _{0}\leq S}{\mbox{argmin}}\Vert\mathbf{Z}-\mathbf{X}^{k}\Vert_{F}^{2},
\end{equation}

\noindent to which the solution is a hard thresholded version of $\mathbf{X}^{k}$
given explicitly as $\mathbf{Z}^{k}=\mathcal{H}_{S}\left(\mathbf{X}^{k}\right)$.
Taking both updates together, the complete version of GHT-QPM is given
in Algorithm \ref{alg:GHTA-ALM}. 

\begin{algorithm}[!th]
\textbf{Inputs:} Dictionary $\mathbf{D}$, signal concatenation matrix
$\mathbf{Y}$, penalty parameter $\rho>0$, global budget $S$, and initialization for $\mathbf{Z}^{0}$

\textbf{Output:} Final coefficient matrix $\mathbf{Z}^{k}$

$\,$

$k=1$

Find Cholesky factorization of $\,$$\mathbf{D}^{T}\mathbf{D}+\rho\mathbf{I}\rightarrow\mathbf{C}\mathbf{C}^{T}$

Pre-compute $\mathbf{\mathbf{W}=D}^{T}\mathbf{Y}$

\textbf{While convergence criterion not met}

~~~~~~~Solve $\mathbf{CJ}=\mathbf{\mathbf{W}}+\rho\mathbf{Z}^{k-1}$
for $\mathbf{J}$ via forward substitution

~~~~~~~Solve $\mathbf{C}^{T}\mathbf{X}^{k}=\mathbf{\mathbf{J}}$
for $\mathbf{X}^{k}$ via backward substitution

~~~~~~~Find the projection $\mathbf{Z}^{k}=\mathcal{H}_{S}\left(\mathbf{X}^{k}\right)$

~~~~~~~$k\leftarrow k+1$

\textbf{End}

\caption{\label{alg:GHTA-ALM} GHT-QPM}
\end{algorithm}

\subsection{GHT-ADMM \label{sub:GHT-ADMM}}

We also introduce a second method for approximately solving the joint
problem, which is a heuristic form of the popular Alternating Direction
Method of Multipliers (ADMM). While developed close to a half a century
ago, ADMM and other Lagrange multiplier methods in general have seen
an explosion of recent interest in the machine learning and signal
processing communities \cite{boyd2011distributed,goldstein2009split}.
While classically ADMM has been provably mathematically convergent
for only convex problems, recent work has also proven convergence
of the method for particular families of nonconvex problems (see e.g.,
\cite{zhang2010alternating,xu2012alternating,hong2014convergence,magnusson2014convergence}).
There has also been extensive successful use of ADMM as a heuristic
method for highly nonconvex problems \cite{xu2012alternating,zhang2010alternating,Watt,boyd2011distributed,barman2011decomposition,derbinsky2013improved,fu2013bethe,YOU_ADMM}.
It is in this spirit that we have applied ADMM to our nonconvex problem
and, like these works, find it to provide excellent results empirically
(see Section \ref{sec:Experiments}).

To achieve an ADMM algorithm for the joint problem we rewrite it by
again introducing a surrogate variable $\mathbf{Z}$ as \noindent
\begin{equation}
\begin{aligned}\underset{\mathbf{X},\mathbf{Z}}{\mbox{minimize}} & \,\,\Vert\mathbf{D}\mathbf{X}-\mathbf{Y}\Vert_{F}^{2}\\
\mbox{subject to} & \,\,\Vert\mathbf{Z}\Vert_{0}\leq S\\
 & \,\,\mathbf{X}=\mathbf{Z}.
\end{aligned}
\end{equation}
We then form the Augmented Lagrangian associated with this problem,
given by \noindent
\begin{equation}
\begin{array}{c}
\mathcal{L}\left(\mathbf{X},\mathbf{Z},\boldsymbol{\Lambda},\rho\right)=\Vert\mathbf{D}\mathbf{X}-\mathbf{Y}\Vert_{F}^{2}\\
+\rho\Vert\mathbf{X}-\mathbf{Z}\Vert_{F}^{2}+\left\langle \boldsymbol{\Lambda},\,\mathbf{X}-\mathbf{Z}\right\rangle
\end{array}
\end{equation}
where $\boldsymbol{\Lambda}$ is the dual variable, $\left\langle \cdot,\cdot\right\rangle $
returns the inner-product of its input matrices, and $\mathbf{Z}$
is constrained such that $\Vert\mathbf{Z}\Vert_{0}\leq S$. With ADMM
we repeatedly take a single Gauss-Seidel sweep across the primal variables,
minimizing $\mathcal{L}$ independently over $\mathbf{X}$ and $\mathbf{Z}$
respectively, followed by a single dual ascent step in $\boldsymbol{\Lambda}$.
This gives the closed form updates for the two primal variables as\noindent
\begin{equation}
\begin{aligned}\mathbf{X}^{k}=\left(\mathbf{D}^{T}\mathbf{D}+\rho\mathbf{I}\right)^{-1}\left(\mathbf{D}^{T}\mathbf{Y}+\rho\mathbf{Z}^{k-1}-\boldsymbol{\Lambda}^{k-1}\right)\\
\begin{array}{c}
\mathbf{Z}^{k}=\mathcal{H}_{S}\left[\mathbf{X}^{k}+\frac{1}{\rho}\boldsymbol{\Lambda}^{k-1}\right]\end{array}
\end{aligned}
\end{equation}
Again the linear system in the $\mathbf{X}$ update is solved effectively
via catched Cholesky factorization, and $\mathcal{H}_{S}\left(\cdot\right)$
is the hard thresholding operator. The associated dual ascent update
step is then given by\noindent
\begin{equation}
\boldsymbol{\Lambda}^{k}=\boldsymbol{\Lambda}^{k-1}+\rho\left(\mathbf{X}^{k}-\mathbf{Z}^{k}\right).
\end{equation}
For convenience we summarize the ADMM heuristic used in this paper in Algorithm \ref{alg:GHTA-ADMM}.

\begin{algorithm}[!th]
\textbf{Inputs:} Dictionary $\mathbf{D}$, signal concatenation matrix
$\mathbf{Y}$, penalty parameter $\rho>0$, global budget $S$, and initializations for $\mathbf{Z}^{0}$
and $\boldsymbol{\Lambda}^{0}$

\textbf{Output:} Final coefficient matrix $\mathbf{Z}^{k}$

$\:$

$k=1$

Find Cholesky factorization of $\,$$\mathbf{D}^{T}\mathbf{D}+\rho\mathbf{I}\rightarrow\mathbf{C}\mathbf{C}^{T}$

Pre-compute $\mathbf{\mathbf{W}=D}^{T}\mathbf{Y}$

\textbf{While convergence criterion not met}

~~~~~~~Solve $\mathbf{CJ}=\mathbf{\mathbf{W}}+\rho\mathbf{Z}^{k-1}-\boldsymbol{\Lambda}^{k-1}$
for $\mathbf{J}$

~~~~~~~Solve $\mathbf{C}^{T}\mathbf{X}^{k}=\mathbf{\mathbf{J}}$
for $\mathbf{X}^{k}$

~~~~~~~Find the projection $\mathbf{Z}^{k}=\mathcal{H}_{S}\left(\mathbf{X}^{k}+\frac{1}{\rho}\boldsymbol{\Lambda}^{k-1}\right)$

~~~~~~~Update the dual variable $\boldsymbol{\Lambda}^{k}=\boldsymbol{\Lambda}^{k-1}+\rho\left(\mathbf{X}^{k}-\mathbf{Z}^{k}\right)$

~~~~~~~$k\leftarrow k+1$

\textbf{End}

\caption{\label{alg:GHTA-ADMM} GHT-ADMM}
\end{algorithm}

\subsection{Time complexity analysis \label{sub:Complexity-analysis}}

In this Section we derive time complexities of both proposed algorithms. In what follows we assume that i) $L=2N$, that is the dictionary is two times overcomplete, and ii) the number of signals $P$ greatly dominates every other influencing parameter.

As can be seen in Algorithm \ref{alg:GHTA-ALM}, each iteration of GHT-QPM includes i) solving
a linear system of equations to update $\mathbf{X}^{k}$, and ii)
hard thresholding the solution to update $\mathbf{Z}^{k}$. In our
implementation of GHT-QPM we pre-compute $\mathbf{D}^{T}\mathbf{Y}$
as well as the Cholesky factorization of the matrix $\mathbf{D}^{T}\mathbf{D}+\rho\mathbf{I}$
outside the loop and as a result, updating $\mathbf{X}^{k}$ can be
done more cheaply via forward/backward substitutions inside the loop.
Assuming $\mathbf{D}\in\mathbb{R}^{N\times 2N}$ and $\mathbf{Y}\in\mathbb{R}^{N\times P}$,
construction of matrices $\mathbf{D}^{T}\mathbf{Y}$ and $\mathbf{D}^{T}\mathbf{D}+\rho\mathbf{I}$
require $4N^{2}P$ and $4N^{3}+2N$ operations, respectively.
In our analysis we do not account for matrix (re)assignment operations
that can be dealt with memory pre-allocation. Additionally, whenever
possible we can take advantage of the symmetry of the matrices involved,
as is for example the case when computing $\mathbf{D}^{T}\mathbf{D}+\rho\mathbf{I}$.
Finally, considering $\frac{8}{3}N^{3}$ operations required for Cholesky
factorization of $\mathbf{D}^{T}\mathbf{D}+\rho\mathbf{I}$, the outside-the-loop cost of GHT-QPM adds up to $4N^{2}P+\frac{20}{3}N^{3}+2N$, that is $\mathcal{O}(N^{2}P)$.

\noindent Now to compute the per iteration cost of GHT-QPM, the cost
of hard thresholding operation must be added to the $8N^{2}P$ operations
needed for forward and backward substitutions, as well as the $4NP$ operations
required for computing $\mathbf{D}^{T}\mathbf{Y}+\rho\mathbf{Z}^{k}$.
Luckily, we are only interested in finding the $S$ largest (in magnitude)
elements of $\mathbf{X}^{k}$, where $S$ is typically much smaller
than $2NP$ - the total number of elements in $\mathbf{X}^{k}$. A
number of efficient algorithms have been proposed to find the $S$
largest (or smallest) elements in an array, that run in linear time
\cite{hoare1961algorithm,floyd1975algorithm,martinez2004partial}.
In particular Hoare's selection algorithm \cite{hoare1961algorithm},
also known as \textit{quickselect}, runs in $\mathcal{O}(NP)$.
Combined together, the per iteration cost of GHT-QPM adds up to $8N^{2}P+4NP+\mathcal{O}(NP)$, that is again $\mathcal{O}(N^{2}P)$.

The time complexity analysis of GHT-ADMM is essentially similar to
that of GHT-QPM, with a few additional steps: subtraction of $\boldsymbol{\Lambda}^{k-1}$
from $\mathbf{D}^{T}\mathbf{Y}+\rho\mathbf{Z}^{k}$ in updating $\mathbf{X}^{k}$
which takes $2NP$ operations, addition of $\frac{1}{\rho}\boldsymbol{\Lambda}^{k-1}$
to $\mathbf{X}^{k}$ in updating $\mathbf{Z}^{k}$ which requires
$4NP$ more operations, and finally the dual variable update $\boldsymbol{\Lambda}^{k}\leftarrow\boldsymbol{\Lambda}^{k-1}+\rho\left(\mathbf{X}^{k}-\mathbf{Z}^{k}\right)$
which adds $6NP$ operations to the per iteration cost of GHT-ADMM. Despite these additional computations, solving the two linear systems remains the most expensive step, and hence the time complexity of GHT-ADMM is $\mathcal{O}(N^{2}P)$, akin to that of GHT-QPM.

%
%

\section{Experiments}
\label{sec:Experiments}

In this Section we present the results of applying our proposed global
hard thresholding algorithms to several sparse representation and
recovery problems. For both GHT-QPM and GHT-ADMM and for all synthetic and real experiments we
kept $\rho$ fixed at $\rho=0.1$, however we found that both algorithms are fairly robust to the choice of this parameter. We also initialized both \textbf{$\mathbf{Z}$}
and $\mathbf{\boldsymbol{\Lambda}}$ as zero matrices. As a stopping condition, we ran both algorithms until subsequent differences of the RMSE value $\sqrt{\frac{\left\Vert \mathbf{D}\mathbf{Z}^{k}-\mathbf{Y}\right\Vert _{F}^{2}}{P}}$, where $P$ is again the total number of patches, was less than $10^{-5}$. In all experiments we compare our approach with popular approaches that work on the patch
level: the Orthogonal Matching Pursuit (OMP) algorithm as implemented
in the SparseLab package \cite{donoho2007sparselab}, the Accelerated
Iterative Hard Thresholding (AIHT) algorithm \cite{blumensath2012accelerated},
and the Compressive Sampling Matching Pursuit (CoSaMP) algorithm \cite{needell2009cosamp}.
All experiments were run in MATLAB R2012b on a machine with a 3.40
GHz Intel Core i7 processor and 16 GB of RAM.

\subsection{Experiment on synthetic data}

\begin{figure*}[!th]
\noindent \begin{centering}
\includegraphics[scale=0.55]{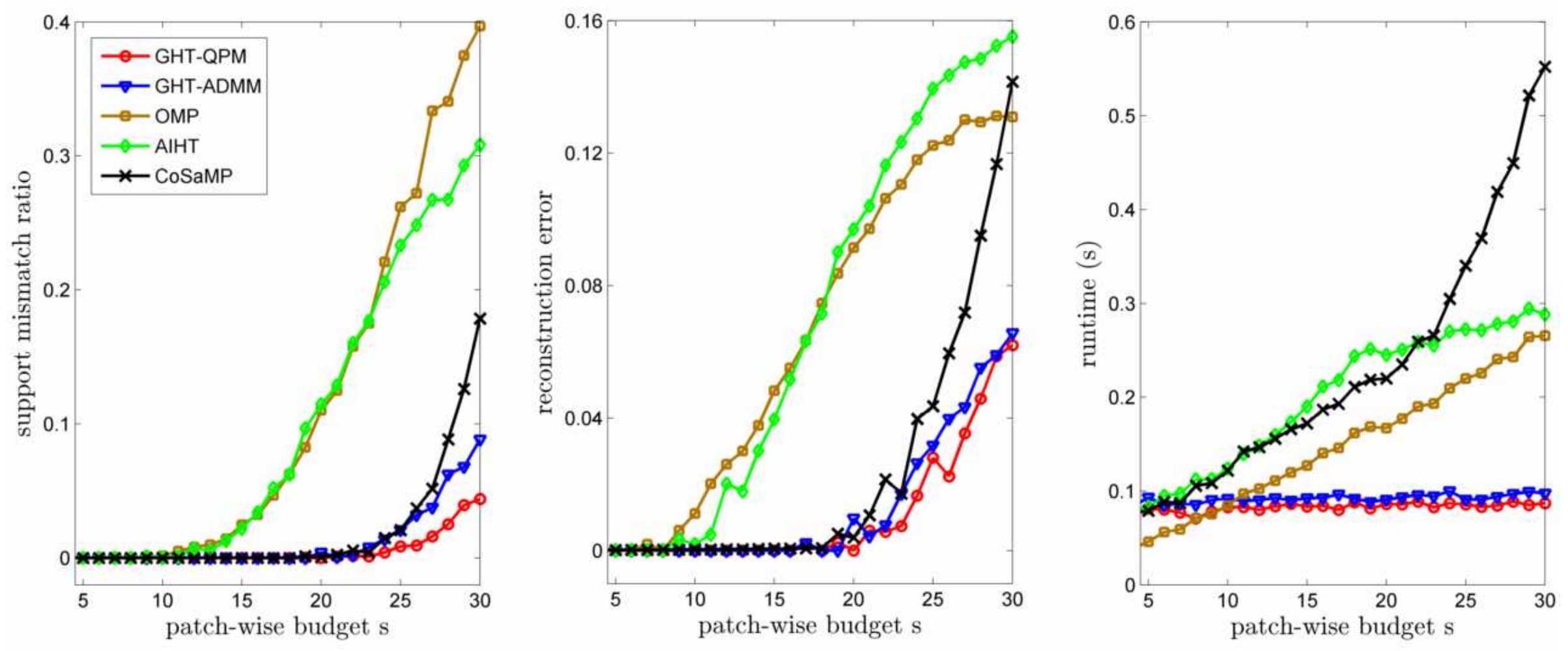}
\par\end{centering}

\caption{\label{fig:Synthetic experiment}Comparison of our methods with patch-wise
greedy and hard thresholding approaches on synthetic data, in terms
of (left) support mismatch ratio, (middle) reconstruction error, and
(right) computation time. The results are plotted for $s=5...30$.}
\end{figure*}

We begin with a simple synthetic experiment where we create an overcomplete
matrix $\mathbf{D}$ of size $100\times200$ whose entries are generated
from a Gaussian distribution (with zero mean and standard deviation of $0.1$). We then generate $P=100$ $s$-sparse
signals $\mathbf{x}_{p}$ for $p=1...100$ each consisting of $s$
nonzero entries taking on the values $\pm1$ uniformly. We then set
$\mathbf{y}_{p}=\mathbf{D}\mathbf{x}_{p}$ for all $p$ and either
solve $100$ instances of the local problem according to the model
in (\ref{eq:Standard L-0 problem}) using a patch-wise competitor
algorithm, or by our methods using the joint model in (\ref{eq: Simultaneous L0 problem})
where the global budget $S$ is set to $100s$. This procedure is
repeated for each value of $s$ in the range of $s=5...30$ and with
$5$ different dictionaries generated as described above. Finally
the average support mismatch ratio, reconstruction error, and computation
time are reported and displayed in Figure \ref{fig:Synthetic experiment}.
Of these three criteria, the first two measure how close the recovered
solution is to the true solution while the last one captures each
algorithm's runtime and how it varies by changing the sparsity level
$s$. More specifically, support mismatch ratio measures the distance
between the support%
\footnote{The support of a matrix is defined as the set of indices with nonzero
values.%
} of the true solution $\mathbf{X}$ denoted by $\mathcal{A}$ and
that of the recovered solution $\mathbf{Z}^{K}$ denoted by $\mathcal{B}$,
and is given by \cite{elad2010sparse}\noindent
\begin{multline}
\textrm{mismatch ratio}=\frac{\textrm{max}\left\{ \left|\mathcal{A}\right|,\,\left|\mathcal{B}\right|\right\} -\left|\mathcal{A}\cap\mathcal{B}\right|}{\textrm{max}\left\{ \left|\mathcal{A}\right|,\,\left|\mathcal{B}\right|\right\} }.
\end{multline}
Here $\left|\mathcal{A}\right|=\left|\mathcal{B}\right|=100s$, and
a mismatch ratio of zero indicates perfect recovery of the true support.
Reconstruction error (or RMSE) on the other hand measures how close the recovered
representation $\mathbf{DZ}^{k}$ is to $\mathbf{Y}$.
As can be seen in Figure \ref{fig:Synthetic experiment}, both GHT-QPM
and GHT-ADMM are competitive with the best of the patch-wise algorithms
in all three of these categories. Interestingly, while both global
hard thresholding algorithms match the leading algorithm (CoSaMP)
in terms of support mismatch ratio and reconstruction error, they
also match the algorithm with best computation time (OMP). Therefore
these experiments seem to indicate that global hard thresholding provides
the best of both worlds: algorithms with high accuracy and low computation
time. It is also worth noting that unlike the competitors, the computation
time of both proposed algorithms do not increase as the per patch
sparsity level $s$ increases, as expected from the time complexity
analysis in Subsection \ref{sub:Complexity-analysis} of Section \ref{sec:Global-Hard-Thresholding}. Finally, note
that in this experiment we did not take full advantage of the power
of the joint model in (\ref{eq: Simultaneous L0 problem}) over the
patch-wise model in (\ref{eq:Standard L-0 problem}) since all the
synthetic patches were created using the same number $s$ of dictionary
atoms, and this number was given to all patch-wise algorithms. However
the assumption that all patches taken from natural images could be
well represented using the same number of atoms does not typically
hold, and as we will see next the experiments
on real data show that our global methods \textit{significantly} outperform
patch-wise algorithms in terms of reconstruction error.

\subsection{Sparse representation of megapixel images}

\begin{figure*}[!th]
\noindent \begin{centering}
\includegraphics[scale=0.27]{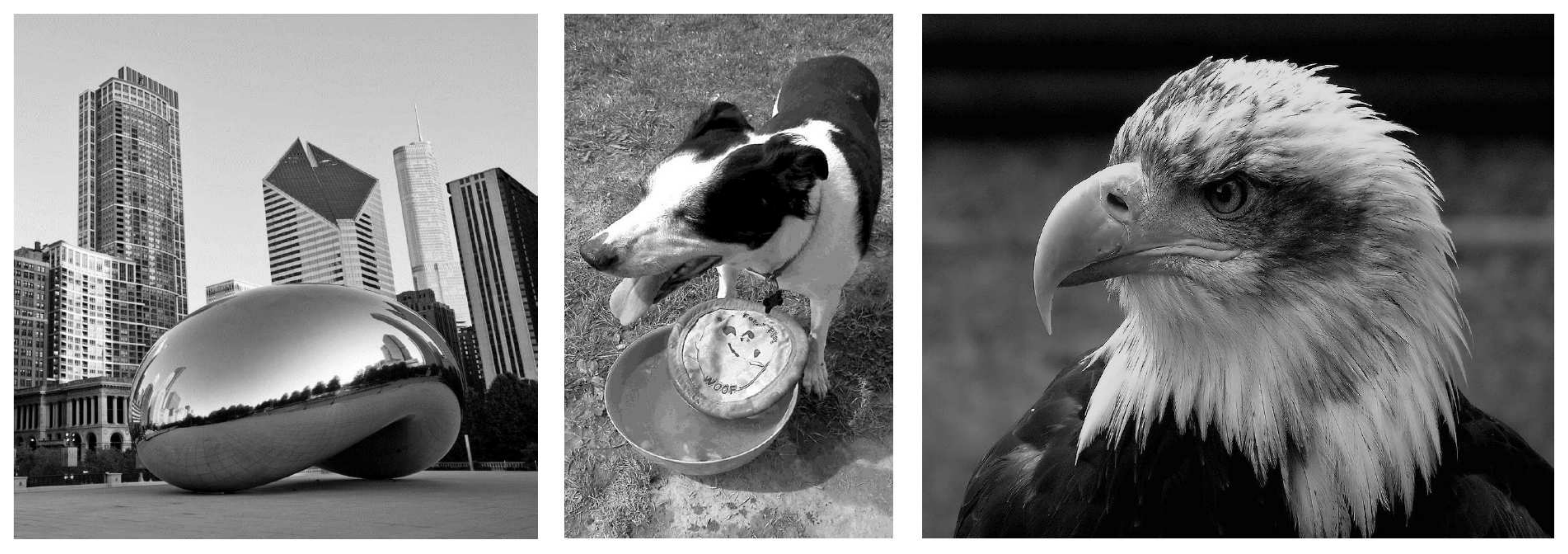}
\par\end{centering}

\caption{\label{fig:sparse rep images}Megapixel gray scale images used for
sparse representation and denoising experiments. From left to right:
Chicago; size: $1024\times1024$, hot-dog; size: $800\times1360$, and eagle; size: $1200\times880$.}
\end{figure*}

Here we perform a series of experiments on sparse representation of
large megapixel images, using those images displayed in Figure \ref{fig:sparse rep images}.
In the first set of experiments we compare GHT-QPM and GHT-ADMM to
OMP, AIHT, and CoSaMP in terms of their ability to sparsely represent
these images. This experiment illustrates the surprising efficacy
and scalability of our global hard thresholding approaches.

\label{ssec:subhead-1}

Decomposing each image into $P$ non-overlapping $8\times8$ patches,
this data is columnized into $64\times1$ vectors and concatenated
into a $64\times P$ matrix referred to as $\mathbf{Y}$. We then
learn sparse representations for these image patches over a $64\times100$
overcomplete DCT dictionary. Specifically, we run each patch-wise
algorithm using an average per patch budget of $s=\nicefrac{S}{P}$
nonzero coefficients for $s$ in a range of $5$ to $30$ in increments
of $1$. We then run both global algorithms on the entire data set
$\mathbf{Y}$ using the global budget $S$.

\begin{figure*}[!th]
\noindent \begin{centering}
\includegraphics[scale=0.52]{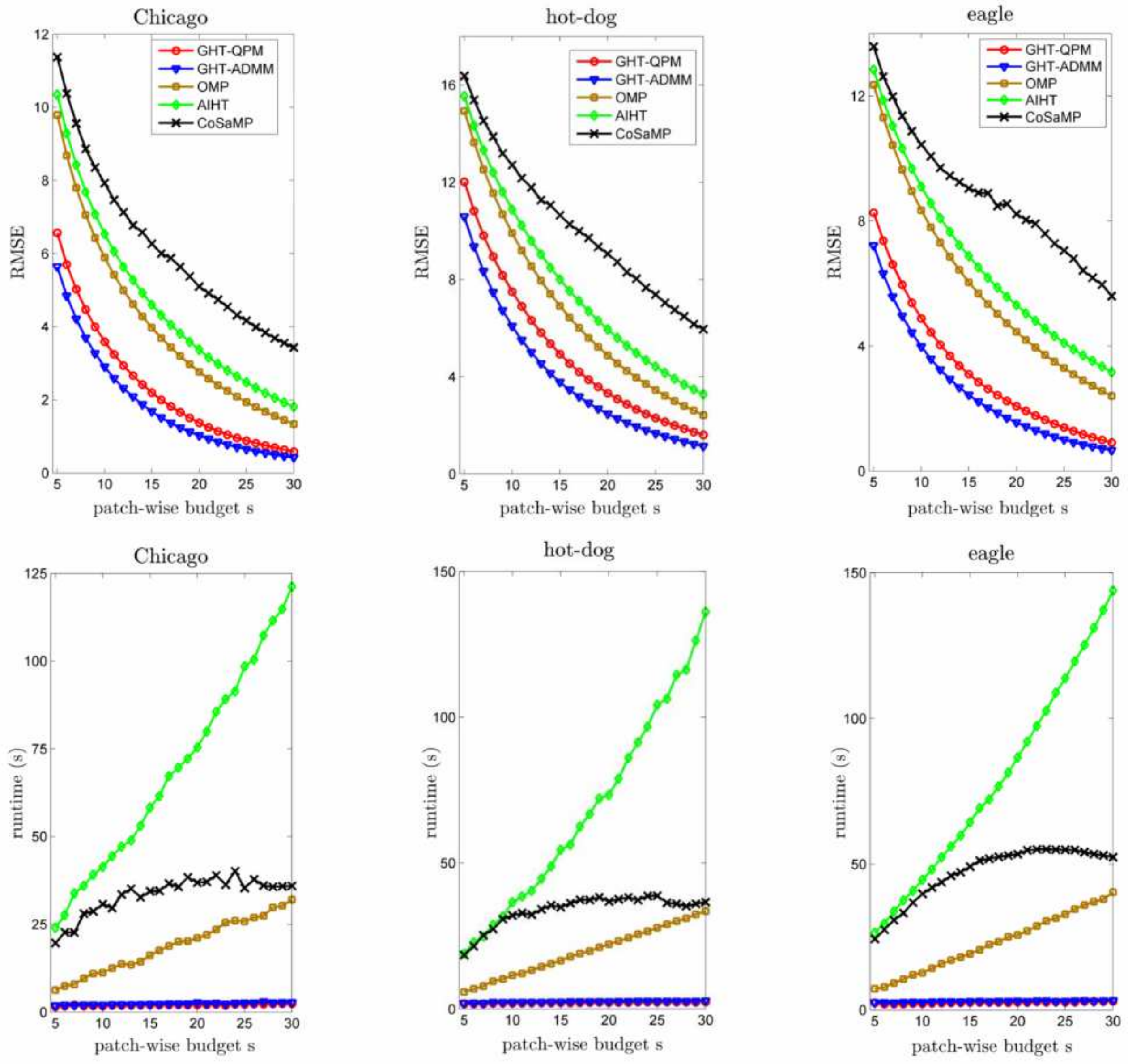}
\par\end{centering}

\caption{\label{fig:sparse-rep experiment}Comparison of our methods with patch-wise
greedy and hard thresholding approaches on real data. (top panels)
RMSE plotted as a function of patch-wise nonzero budget $s$ over
a range of $5$ to $30$. Both GHT-QPM and GHT-ADMM significantly
outperform the nearest competitor in terms of RMSE. (bottom panels)
Computation time for the algorithms compared at each budget level.
While the total runtime of competing algorithms increases substantially
with the budget level, the run times of our algorithms remain stable
across the range of tested budgets. }
\end{figure*}

Figure \ref{fig:sparse-rep experiment} displays the results of these
experiments on the megapixel images from Figure \ref{fig:sparse rep images},
including final root mean squared errors and runtimes of the
associated algorithms. In all instances both of our global methods
significantly outperform the various patch-wise methods in terms of
RMSE over the entire budget range. For example, with a patch-wise
budget of $\nicefrac{S}{P}=10$ the RMSE of our algorithms range between
$78\%$ and $350\%$ lower than the nearest competitor. This major
difference in reconstruction error lends credence to the claim that
GHT algorithms effectively distribute the global budget across all
patches of the megapixel image. While not overly surprising given
the wealth of work on ADMM heuristic algorithms (see Subsection \ref{sub:GHT-ADMM} of Section \ref{sec:Global-Hard-Thresholding}),
it is interesting to note that GHT-ADMM outperforms GHT-QPM (as well
as all other competitors) in terms of RMSE. Moreover, the total computation
time of our algorithms remain fairly stable across the range of budgets
tested, while competitors' runtimes can increase quite steeply as
the nonzero budget is increased.

\begin{figure*}[!th]
\noindent \begin{centering}
\includegraphics[scale=0.4]{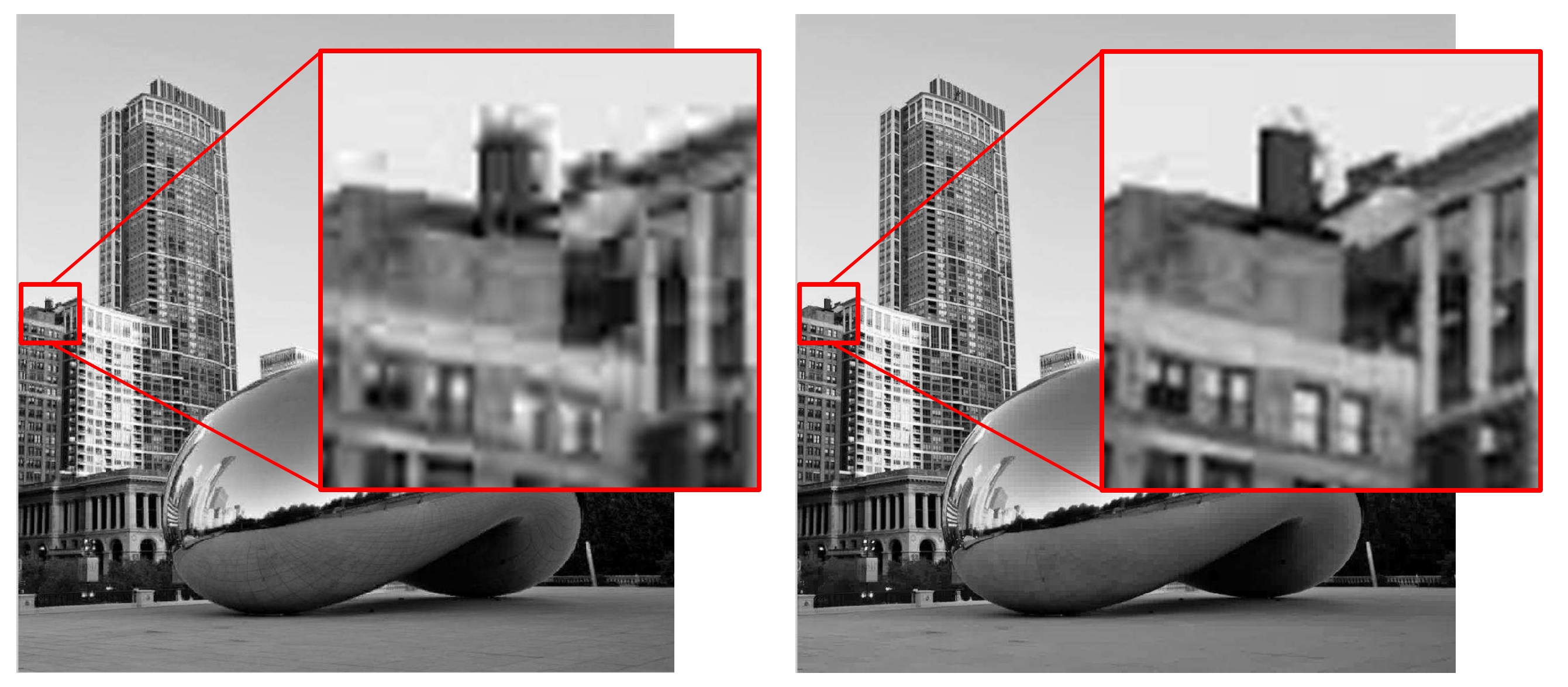}
\par\end{centering}

\caption{\label{fig:sparse-rep-close-up}Close-up visual comparison of OMP
(the best competitor) and GHT-QPM (the inferior of our two proposed
algorithms). OMP (left) uses a per patch budget of $\nicefrac{S}{P}=5$
nonzeros while GHT-QPM (right) is given the equivalent global budget
of $S=5P$. }
\end{figure*}

To compare visual quality of the images reconstructed by different
algorithms, we show in Figure \ref{fig:sparse-rep-close-up} the results
obtained by OMP (the best competitor) and GHT-QPM (the inferior of
our two proposed algorithms) using a total budget of $S=5P$ on one
of test images. The close-up comparison between the two methods clearly
shows the visual advantage gained by solving the joint model.

\label{ssec:subhead}

\label{ssec:subhead-2}

\subsection{Runtime and convergence on a million patches}

\begin{figure}[!th]
\noindent \begin{centering}
\includegraphics[scale=0.4]{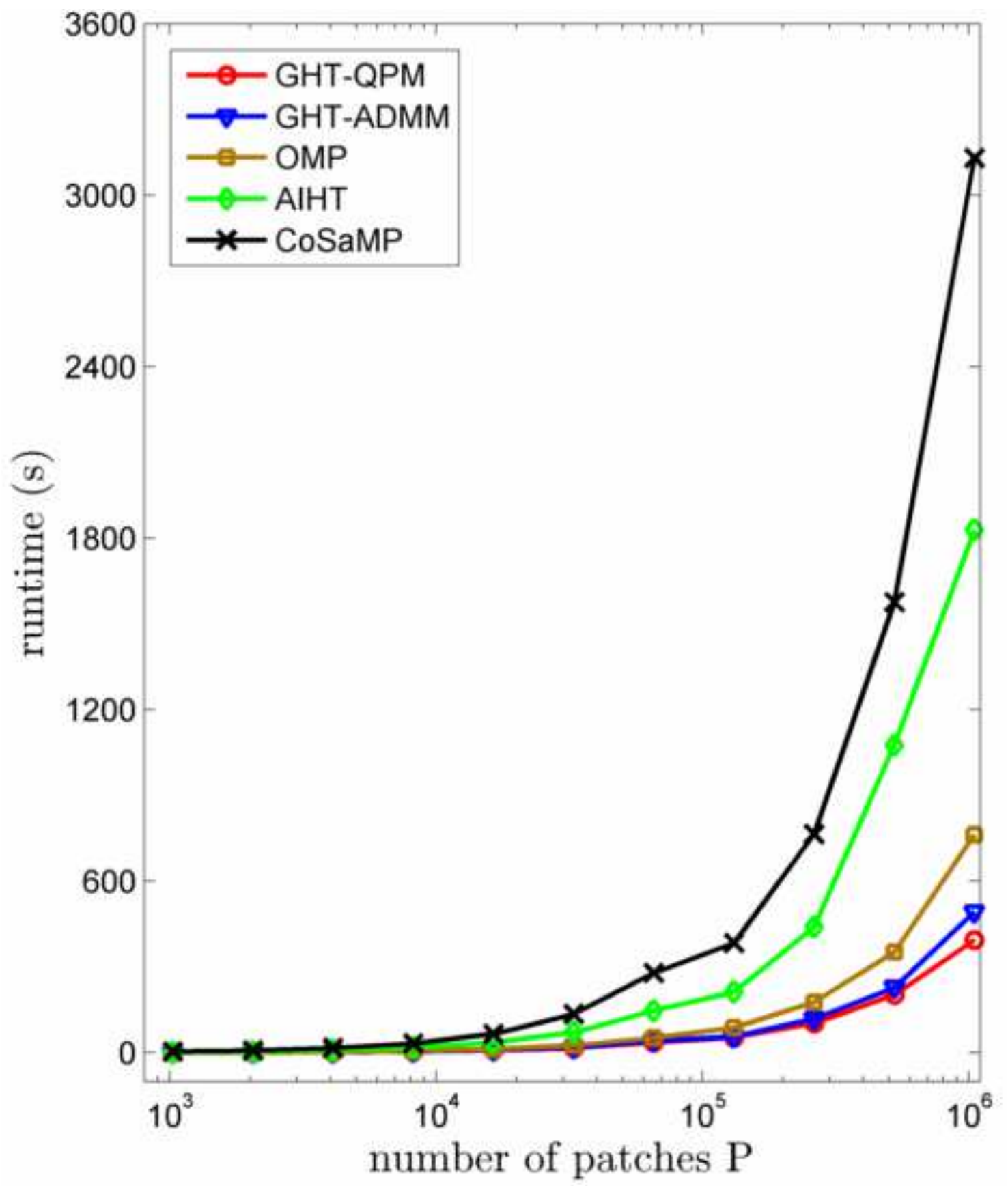}
\par\end{centering}

\caption{\label{fig:Patch_time} Runtime comparison of all patch-wise and global
algorithms computed over a wide range ($2^{10}$ to $2^{20}$) of
randomly selected $8\times8$ patches. }
\end{figure}

So far in both synthetic and real sparse representation experiments
we have kept the number of patches $P$ fixed and only varied the
patch-wise budget $s$. It is also of practical interest to explore
how the runtimes of patch-wise and global algorithms are affected
when the number of patches increase while keeping a fixed patch-wise
budget. To conduct this experiment we collect a set of $P$ random
$8\times8$ patches taken from a collection of natural images \cite{hyvarinen2009natural}.
For each $P\in\left\{ 2^{10},\,2^{11},\,\ldots,2^{20}\right\} $ we
then run all the algorithms keeping a fixed per patch budget of $s=\nicefrac{S}{P}=10$
and plot the computation times in Figure \ref{fig:Patch_time}. As
expected the runtime of both our algorithms are linear in $P$ (note
that the scale on the x-axis is logarithmic). This Figure confirms
that both GHT algorithms are highly scalable. As can be seen GHT-ADMM
runs slightly slower than GHT-QPM which is consistent with our time
complexity analysis.

\begin{figure*}[!th]
\noindent \begin{centering}
\includegraphics[scale=0.50]{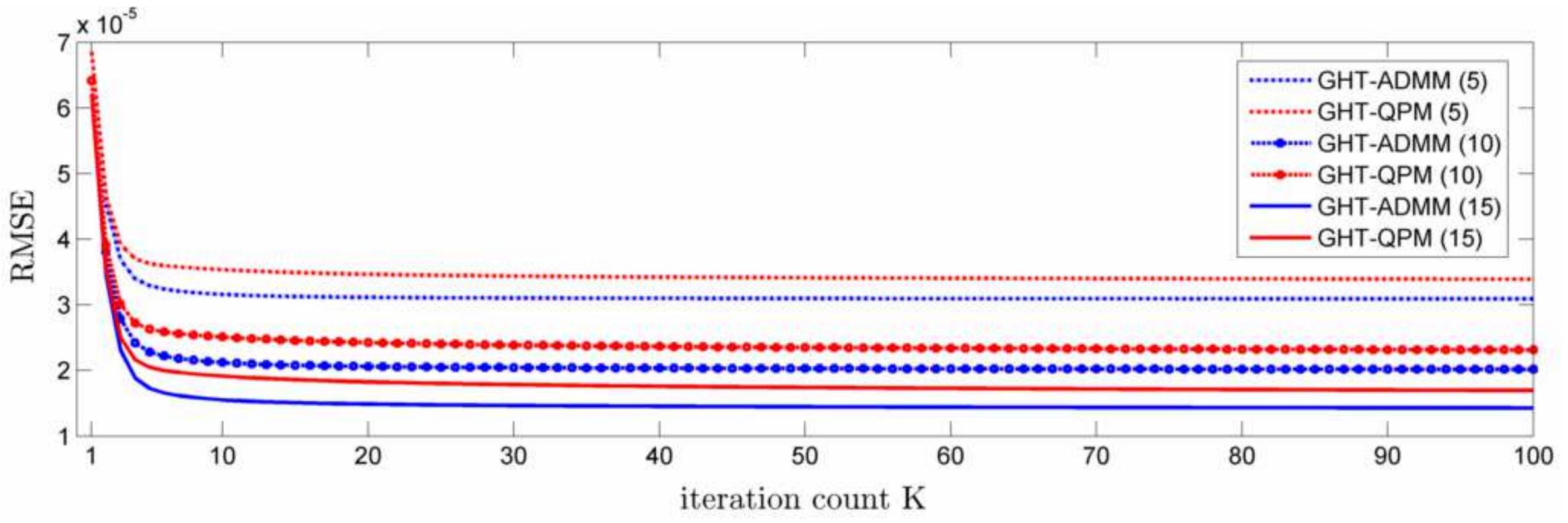}
\par\end{centering}

\caption{\label{fig: GHTA convergence}Change of RMSE over iteration count
for GHT-QPM and GHT-ADMM. Three global budget levels are used: $S=5P$,
$10P$, and $15P$.}
\end{figure*}

Finally, we choose three global sparsity budgets such that an average
of $s=5$, $10$, and $15$ atoms would be used per individual patch
and run GHT-QPM and GHT-ADMM to resolve $10^{6}$ randomly selected
patches. In Figure \ref{fig: GHTA convergence} we plot for each iteration
$k=1...100$ the RMSE value. Note three observations: firstly, both algorithms have decreasing
values of $\left\Vert \mathbf{D}\mathbf{Z}^{k}-\mathbf{Y}\right\Vert $
empirically\footnote{Denoting by $f(\mathbf{X},\mathbf{Z})=\left\Vert \mathbf{D}\mathbf{X}-\mathbf{Y}\right\Vert _{F}^{2}+\rho\left\Vert \mathbf{X}-\mathbf{Z}\right\Vert _{F}^{2}$
as well as $\mathbf{X}^{k}=\underset{\mathbf{X}}{\textrm{argmin}}f(\mathbf{X},\mathbf{Z}^{k-1})$
for some $\mathbf{Z}^{k-1}$ and $\mathbf{Z}^{k}=\underset{\left\Vert \mathbf{Z}\right\Vert _{0}\leq S}{\textrm{argmin}}f(\mathbf{X}^{k},\mathbf{Z})$,
then it follows that $f(\mathbf{X}^{k},\mathbf{Z}^{k})\leq f(\mathbf{X}^{k},\mathbf{Z}^{k-1})\leq f(\mathbf{X}^{k-1},\mathbf{Z}^{k-1})$.
Hence the QPM approach produces iterates $\left\{ \mathbf{X}^{k},\mathbf{Z}^{k}\right\} $
that are non-increasing in the objective.}, secondly that GHT-ADMM gives lower reconstruction errors
compared to GHT-QPM across all budget levels, and third that within as few as just $10$ iterations both algorithms have converged.

\subsection{Natural image denoising}

\begin{figure*}[!th]
\noindent \begin{centering}
\includegraphics[scale=.8]{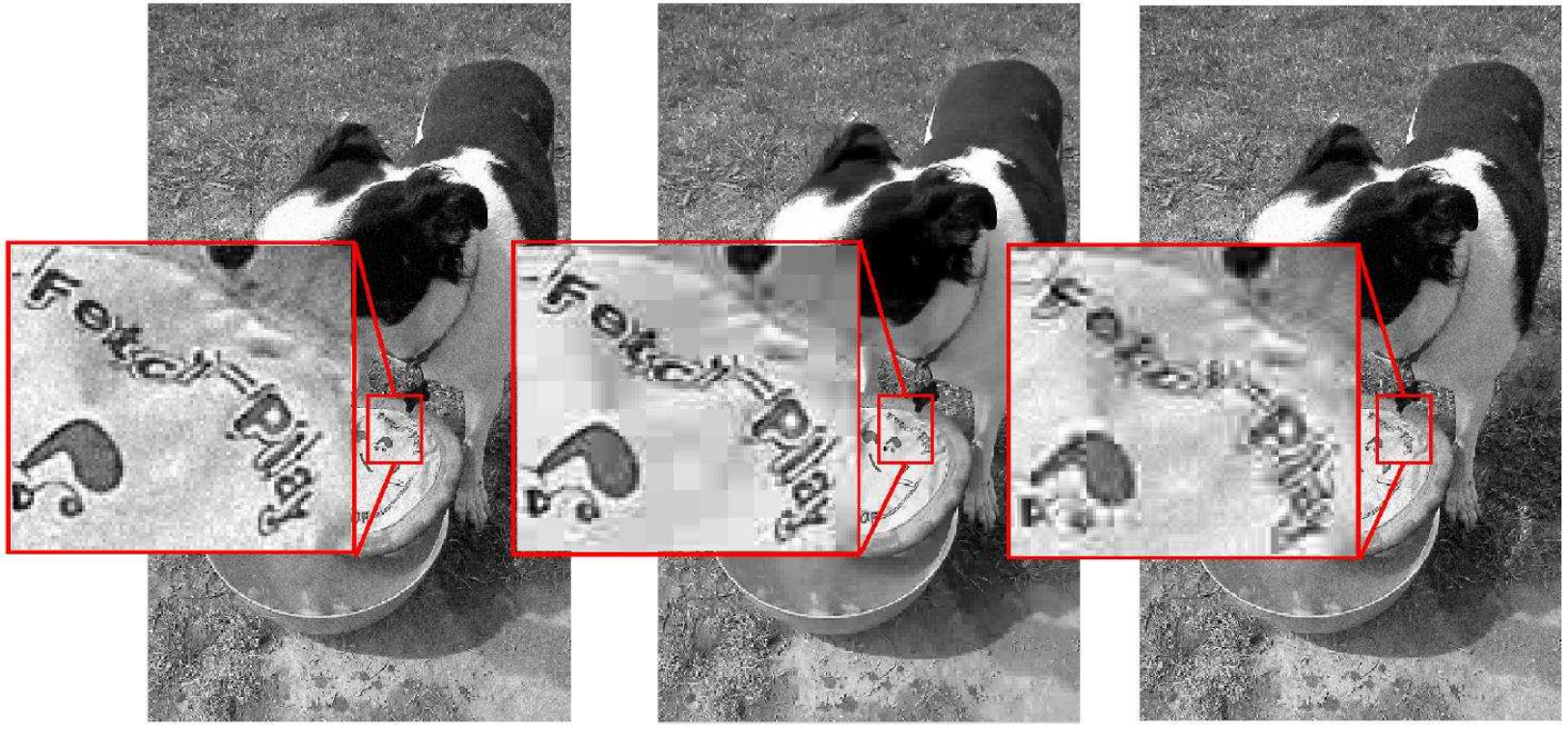}
\par\end{centering}

\caption{\label{fig:Nurgetson}Close-up visual comparison of OMP and GHT-ADMM for sparse image denoising. (left) Noisy image with $\sigma=10$. OMP (right) uses a per patch budget of $s=10$ nonzeros while GHT-ADMM (middle) is given the equivalent global budget of $S=10P$.}
\end{figure*}

In this next set of experiments, we add Gaussian noise to the images
in Figure \ref{fig:sparse rep images} and test the efficacy of our
proposed methods for noise removal. More specifically, for a range
of noise levels $\sigma=5$, $10$, $20$, $30$, and $40$ we add
zero mean Gaussian noise with standard deviation $\sigma$ to these
images, which are then as before decomposed into $P$ non-overlapping
$8\times8$ patches, each of which is columnized, and a $64\times P$
matrix $\mathbf{Y}$ is formed. Both of our global algorithms, GHT-QPM
and GHT-ADMM, are then given this entire matrix to denoise along with
a global budget of $S=10P$ nonzero coefficients. The competing algorithms
are given the analogous patch-wise budgets of $\nicefrac{S}{P}=10$
nonzero coefficients, and process the image on the patch level. Noise
removal results in terms of PSNR (in dB) are tabulated in Tables \ref{tab:d1} through \ref{tab:d4}. For each noise level, PSNR of the best algorithm is boxed.
Here we can see that both of our global methods significantly outperform
patch-wise algorithms, particularly at low to moderate noise levels.
For example, at a noise level of $\sigma=10$ our algorithms greatly
outperform the nearest competitor on the images tested by $2.7$ to
$4.9$ dB, and at $\sigma=15$ both global algorithms produce results
at $1$ to $2.6$ dB lower than the nearest competitor. Finally, in Figure \ref{fig:Nurgetson} we show an example of a megapixel image from Figure \ref{fig:sparse rep images} to which this amount of noise has been added, and the result of applying GHTA-ADMM (PSNR=26.91 dB) as well as OMP (PSNR=24.12 dB) to denoising the image. As can be seen, the resulting denoised image using GHTA-ADMM is also visually superior, recovering high frequency portions of the image with greater accuracy than OMP.

\begin{table}[htbp]
\centering
\caption{\bf Denoising results (PSNR) for the Chicago image}
\begin{tabular}{cccccc}
\hline
Algorithm & $\sigma=5$ & $\sigma=10$ & $\sigma=20$ & $\sigma=30$ & $\sigma=40$ \\
\hline
OMP & 28.01 & 27.14 & 24.61 & 22.14 & 20.02 \\
AIHT & 27.59 & 26.85 & 24.62 & 22.29 & 20.22 \\
CoSaMP & 26.81 & 26.14 & 24.04 & 21.74 & 19.67 \\
GHT-QPM & 31.49 & 30.54 & \fbox{26.51} & \fbox{23.01} & \fbox{20.41} \\
GHT-ADMM & \fbox{32.61} & \fbox{31.01} & 26.20 & 22.77 & 20.28 \\
\hline
\end{tabular}
  \label{tab:d1}
\end{table}

\begin{table}[htbp]
\centering
\caption{\bf Denoising results (PSNR) for the hot-dog image}
\begin{tabular}{cccccc}
\hline
Algorithm & $\sigma=5$ & $\sigma=10$ & $\sigma=20$ & $\sigma=30$ & $\sigma=40$ \\
\hline
OMP & 24.52 & 24.12 & 22.67 & 20.89 & 19.20 \\
AIHT & 24.21 & 23.87 & 22.59 & 20.93 & 19.29 \\
CoSaMP & 23.73 & 23.39 & 22.15 & 20.49 & 18.85 \\
GHT-QPM & 26.41 & 26.07 & 24.46 & \fbox{21.98} & \fbox{19.74} \\
GHT-ADMM & \fbox{27.45} & \fbox{26.91} & \fbox{24.66} & 21.95 & 19.70 \\
\hline
\end{tabular}
  \label{tab:d3}
\end{table}

\begin{table}[htbp]
\centering
\caption{\bf Denoising results (PSNR) for the eagle image}
\begin{tabular}{cccccc}
\hline
Algorithm & $\sigma=5$ & $\sigma=10$ & $\sigma=20$ & $\sigma=30$ & $\sigma=40$ \\
\hline
OMP & 26.08 & 25.46 & 23.52 & 21.42 & 19.54 \\
AIHT & 25.77 & 25.23 & 23.46 & 21.47 & 19.64 \\
CoSaMP & 25.29 & 24.82 & 23.05 & 21.04 & 19.23 \\
GHT-QPM & 29.55 & 28.87 & \fbox{25.75} & \fbox{22.50} & \fbox{20.04} \\
GHT-ADMM & \fbox{30.61} & \fbox{29.52} & 25.55 & 22.33 & 19.96 \\
\hline
\end{tabular}
  \label{tab:d4}
\end{table}

\section{Conclusions}
\label{sec:Conclusions}

In this work we have described two hard thresholding algorithms for
approximately solving the joint sparse representation problem in (\ref{eq: Simultaneous L0 problem}).
Both are penalty method approaches based on the notion of variable
splitting, with the former being an instance of the Quadratic Penalty Method. While the latter, a heuristic adaptation
of the popular ADMM framework, is not provably convergent it nonetheless
consistently outperforms most of the popular algorithms for sparse
representation and recovery in our extensive experiments on synthetic
and natural image data. These experiments show that our algorithms
distribute a global budget of nonzero coefficients much more effectively
than naive patch-wise methods that use fixed local budgets. Additionally,
both proposed algorithms are highly scalable making them attractive
for researchers working on sparse recovery problems in signal and
image processing. While we have presented experimental results for sparse image representation and denoising, the approaches discussed in this paper
for solving (\ref{eq: Simultaneous L0 problem}) can be applied to a number of further image processing tasks such as image inpainting, deblurring, and super-resolution.



\section{Acknowledgements}

This work is supported
in part by the GK-12 Reach For the Stars program through the National
Science Foundation grant DGE-0948017, and the Department of Energy
grant DE-NA0000457.

\bibliographystyle{elsarticle-num}
\bibliography{GHT}

\end{document}